\documentclass[11pt]{article}

\usepackage{arabtex}
\usepackage{times}
\usepackage{eacl2017}
\usepackage{times}
\usepackage{url}
\usepackage{latexsym}
\usepackage{graphicx}
\graphicspath{.}

\eaclfinalcopy 

\title{SMARTies: Sentiment Models for Arabic Target entities}

\author{Noura Farra  {\normalfont and} Kathleen McKeown \\
  Columbia University \\
  Department of Computer Science \\
  {\tt \{noura,kathy\}@cs.columbia.edu} } 

\begin{document}
\maketitle
\begin{abstract}
 We consider entity-level sentiment analysis in Arabic, a morphologically rich language with increasing resources. We present a system that is applied to complex posts written in response to Arabic newspaper articles.   Our goal is to identify important entity ``targets" within the post along with the polarity expressed about each target. We achieve significant improvements over multiple baselines,
  demonstrating that the use of specific morphological representations improves the performance of identifying both important targets and their sentiment, and that the use of distributional semantic clusters further boosts performances for these representations, especially when richer linguistic resources are not available. 
\end{abstract}

\section{Introduction\label{intro}}

\setarab
\novocalize
Target-specific sentiment analysis has recently become a popular problem in natural language processing.
In interpreting social media posts, analysis needs to include more than just whether people feel positively or negatively; it also needs to include {\em what} they like or dislike.
The task of finding all targets within the data has been called \textbf{``open-domain targeted sentiment''} \cite{mitchell2013open,zhang2015neural}. If we could successfully identify the targets of sentiment, it would be valuable for a number of applications 
including sentiment summarization, question answering, understanding public opinion during political conflict, or assessing needs of populations during natural disasters.

\begin{figure}
	\centering
	\includegraphics[width=0.45\textwidth]{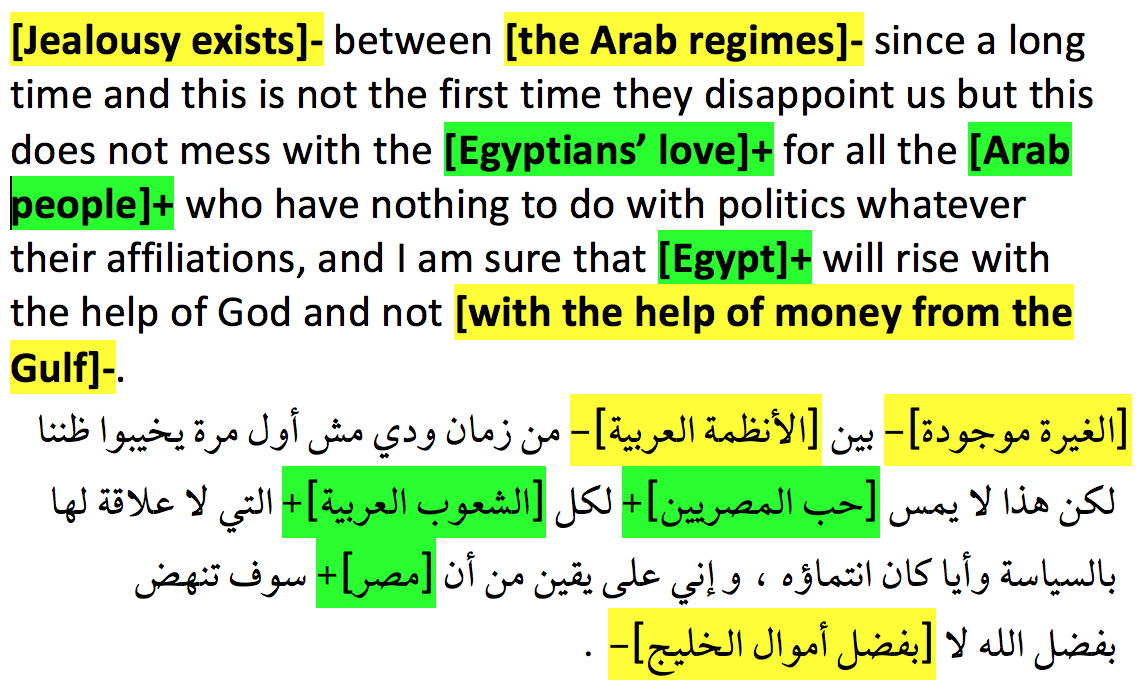}
	\caption{\label{fig1} Online post with annotated target entities and sentiment (green:pos, yellow:neg).}
\end{figure}

In this paper, we address the open-domain targeted sentiment task. Input to our system consists of online posts, which can be comprised of one or multiple sentences, contain multiple entities with different sentiment, and have different domains. Our goal is to identify the important entities towards which opinions are expressed in the post; these can include any nominal or noun phrase, including events, or concepts, and they are not restricted to named entities as has been the case in some previous work. The only constraint is that the entities need to be explicitly mentioned in the text. 
Our work also differs from much work on targeted sentiment analysis in that posts are long, complex, with many annotated targets and a lack of punctuation that is characteristic of Arabic online language. Figure \ref{fig1} shows an example post, where targets are either labeled positive (green) if a positive opinion is expressed about them and negative (yellow) if a negative opinion is expressed.

To identify targets and sentiment, we develop two sequence labeling models, a target-specific model and a sentiment-specific model. Our models try to learn syntactic relations between entities and opinion words, but they also make use of (1) Arabic morphology and (2) entity semantics. Our use of morphology allows us to capture all ``words'' that play a role in identification of the target, while our use of entity semantics allows us to group together similar entities which may all be targets of the same sentiment; for example, if a commenter expresses negative sentiment towards the United States, they may also express negative sentiment towards America or Obama.


Our results show that morphology matters when identifying entity targets and the sentiment expressed towards them. We find for instance that the attaching Arabic definite article \textit{Al+} <Al>  is an important indicator of the presence of a target entity and splitting it off boosts recall of targets, while sentiment models perform better when less tokens are split. We also conduct a detailed analysis of errors revealing that the task generally entails hard problems such as a considerable amount of implicit sentiment and the presence of multiple targets with varying importance. 

In what follows, we describe related work (section \ref{related}), data and models (sections \ref{data} and \ref{system}), and linguistic decisions made for Arabic (section \ref{arabic}). In section \ref{clusters}, we describe our use of word vector clusters learned on a large Arabic corpus. Finally, section  \ref{experiments} presents experiments and detailed error analysis.

\section{Related Work\label{related}}

\paragraph{Aspect-based and Entity-specific  Analysis}
Early work in target-based sentiment looked at identifying aspects in a restricted domain: product or customer reviews. Many of these systems used unsupervised and topical methods for determining aspects of products; \newcite{hu2004mining} used frequent feature mining to find noun phrase aspects, \newcite{brody2010unsupervised} used topic modeling to find important keywords in restaurant reviews, and \newcite{somasundaran2009recognizing} mined the web to find important aspects associated with debate topics and their corresponding polarities.  SemEval 2014 Task 4 \cite{pontiki2014semeval} ran several subtasks for identifying aspect terms and sentiment towards aspects and terms in restaurant and laptop reviews.

Entity-specific sentiment analysis has been frequently studied in social media and online posts. \newcite{jiang2011target} proposed identifying sentiment of a tweet towards a specific named entity, taking into account multiple mentions of the given entity.  \newcite{biyani2015entity} studied sentiment towards entities in online posts, where the local part of the post that contained the entity or mentions of it was identified and the sentiment was classified using a number of linguistic features. The entities were selected beforehand and consisted of known, named entities. More recent work uses LSTM and RNN networks to determine sentiment toward aspects in product reviews \cite{wang-EtAl:2016:EMNLP20163} and towards entities in Twitter \cite{dong2014adaptive,tang2015effective}. SemEval 2016 ran two tasks on sentiment analysis \cite{nakov2016semeval} and stance \cite{mohammad2016semeval} towards pre-defined topics in Twitter, both on English data.

\paragraph{Open domain targeted analysis}
In early work. \newcite{kim2006extracting} proposed finding opinion target and sources
 in news text by automatic labeling of semantic roles. Here, opinion-target relationships were restricted to relations that can be captured using semantic roles. \newcite{ruppenhofer2008finding} discussed the challenges of identifying targets in open-domain text which cannot be addressed by semantic role labeling, such as implicitly conveyed sentiment, global and local targets related to the same entity, and the need for distinguishing between entity and proposition targets.
 
Sequence labeling models became more popular for this problem: \newcite{mitchell2013open}  used CRF model combinations to identify named entity targets in English and Spanish, and \newcite{yang2013joint} used joint modeling to predict opinion expressions and their source and target spans in news articles, improving over several single CRF models. 
Their focus was on identifying directly subjective opinion expressions (e.g "I \textit{hate} [this dictator]"  vs. "[This dictator] is \textit{destroying} his country.")  
Recent work \cite{deng2015joint} identifies entity sources and targets, as well as the sentiment expressed by and towards these entities. This work was based on probablistic soft logic models, also with a focus on direct subjective expressions.    

There is also complementary work on using neural networks for tagging open-domain targets \cite{zhang2015neural,liu2015fine} in shorter posts. Previous work listed did not consider word morphology, or explicitly model distributional entity semantics as indicative of the presence of sentiment targets.

\paragraph{Related work in Arabic}

Past work in Arabic machine translation \cite{habash2006arabic} and named entity recognition \cite{benajiba2008arabic} considered the tokenization of  complex Arabic words as we do in our sequence labeling task. Analysis of such segmentation schemes has not been reported for Arabic sentiment tasks, which cover mostly sentence-level sentiment analysis and where the lemma or surface bag-of-word representations have typically been sufficient.

There are now many studies on sentence-level sentiment analysis in Arabic news and social media \cite{abdul2011subjectivity,mourad2013subjectivity,refaee2014subjectivity,salameh2015sentiment}.  \newcite{elarnaoty2012machine} proposed identifying sources of opinions in Arabic using a CRF with a number of patterns, lexical and subjectivity clues; they did not discuss morphology or syntactic relations. \newcite{al2015human} developed a dataset and built a majority baseline for finding targets in Arabic book reviews of known aspects; \newcite{obaidat2015enhancing} also developed a lexicon-based approach to improve on this baseline. 
\newcite{abu2013identifying} created a simple opinion-target system for Arabic by identifying noun phrases in polarized text; this was done intrinsically as part of an effort to identify opinion subgroups in online discussions. There are no other sentiment target studies in Arabic that we know of. In our experiments, we compare to methods similar to these baseline systems, as well as to results of English work that is comparable to ours. 

\paragraph{Entity Clusters}

It has been shown consistently that semantic word clusters improve the performance of named entity recognition \cite{tackstrom2012cross,zirikly2cross,turian2010word} and semantic parsing \cite{saleh2014semantic}; we are not aware of such work for identifying entity targets of sentiment.

\section{Data\label{data}}

We use the Arabic Opinion Target dataset developed by \newcite{farra2015annotating}, which is publicly available\footnote{www.cs.columbia.edu/\textasciitilde{}noura/Resources.html}.  The data consists of 1177 online comments posted in response to \textit{Aljazeera} Arabic newspaper articles and is part of the Qatar Arabic Language Bank (QALB) corpus \cite{HABASH13.ARC,ZAGHOUANI14.956.L14-1721}. The comments are 1-3 sentences long with an average length of 51 words. They were selected such that they included topics from three domains: politics, culture, and sports. 

Targets are always noun phrases and they are either labeled positive if a positive  opinion is expressed \textit{about} them 
and negative if a negative opinion is expressed (as shown in Figure \ref{fig1}). Targets were identified using an incremental process where first \textit{important} entities were identified, and then entities agreed to be neutral were discarded (the annotation does not distinguish between \textit{neutral} and \textit{subjective neutral}).

The data also contains ambiguous or `undetermined' targets where annotators did agree they were targets, but did not agree on the polarity. We use these targets for training our target model, 
but discard them when training our sentiment polarity model. There are 4886 targets distributed as follows: 38.2\% positive, 50.5\% negative, and 11.3\% ambiguous.
We divide the dataset into a training set (80\%), development set (10\%), and blind test set (10\%), all of which represent the three different domains. 
We make the splits available for researchers to run comparative experiments.



\section{Sequence Labeling Models\label{system}}

For modeling the data, we choose Conditional Random Fields (CRF) \cite{lafferty2001conditional} for the ability to engineer Arabic linguistic features and because of the success of CRF models in the past for entity identification and classification related tasks.  

We build two linear chain CRF models: 

\begin{table}
	\begin{center}
			\setlength{\tabcolsep}{3.5pt}
			\begin{tabular}{ l  l  l l l l }
				\bf  The & \bf dictator & \bf  is & \bf destroying & \bf his & \bf country \\  
				T & T & O & O & O & O  \\ 
				N & N & $\emptyset$ & $\emptyset$ & $\emptyset$ & $\emptyset$ \\ 
			\end{tabular}
	\end{center}
	\caption{\label{table1} {Example of CRF annotations.  } }
\end{table}

\begin{enumerate}
	\item \textbf{Target Model} This model predicts a sequence of labels $\vec{E}$ for a sequence of input tokens $\vec{x}$,
	where \[ E_i  \in \{T (target), O (not\_target)\} \] and each token $x_i$ is represented by a feature vector $\vec{{f_i}_t}$. A token is labeled $T$ if it is part of a target; a target can contain one or more consecutive tokens.
	
	\item \textbf{Sentiment Model} This model predicts a sequence of labels $\vec{S}$ for the sequence $\vec{x}$,
	\[ S_i  \in \{P (pos),N (neg),\emptyset(neutral)\} \] 
	and each token $x_i$ is represented by a feature vector:  
	\[ (\vec{{f_i}_s} , E_i)  ;  E_i  \in \{T , O\} \] 
	
	
	Additionally, this model has the constraint: \[ if  E_i  = T  , S_i  \in \{P,N\}  \] and otherwise \[ S_i = \emptyset \]
	
	\end {enumerate}
	
	The last constraint indicating that sentiment is either positive or negative is ensured by the training data, where we have no examples of target tokens having neutral sentiment.
	The two models are trained independently. Thus, if target words are already available for the data, the sentiment model can be run without training or running the target model. Otherwise, the sentiment model can be run on the output of the target predictor. The sentiment model uses knowledge of whether a word is a target and utilizes context from neighboring words
	whereby the entire sequence is optimized to predict sentiment polarities for the targets. An example sequence is shown in Table \ref{table1}, where \textit{the dictator} is an entity target towards which the writer implicitly expresses negative sentiment.

\section{Arabic Morphology and Linguistics\label{arabic}}
	
	\subsection{Arabic Morphology}
	In Arabic, clitics and affixes can attach to the beginning and end of the word stem, making words complex.  For example, in the sentence \textit{<fAstqblwhA>} \textit{`So they welcomed her'},  the discourse conjuction (\textit{so} +<f>), the opinion target (\textit{her}  <hA>+), opinion holder (\textit{they}  <+uw>), and the opinion expression itself (\textit{welcomed <Astqbl>}) are all collapsed in the same word.
	
	Clitics, such as conjunctions +<w> \textit{w+}, prepositions +<b> \textit{b+}, the definite article <+Al> \textit{Al+} \textit{`the'}  (all of which attach at the beginning), and possessive pronouns and object pronouns <h>+ \textit{+h} <hA>+ \textit{+hA} \textit{`his/her'} or \textit{`him/her'} (which attach at the end) can all function as individual words. Thus, they can be represented as separate tokens in the CRF.
	
	The morphological analyzer MADAMIRA \cite{pasha2014madamira} enables the tokenization of a word using multiple schemes. We consider the following two schemes:
	\begin{itemize}
		\item \textbf{D3}: the Declitization scheme which splits off conjunction clitics, particles and prepositions, \textit{Al+}, and all the enclitics at the end.
		\vspace{-0.3cm}
		\item \textbf{ATB}: the Penn Arabic Treebank tokenization, which separates all clitics above except the definite article \textit{Al+}, which it keeps attached.
	\end{itemize}
	For a detailed description of Arabic concatenative morphology and tokenization schemes, the reader is referred to \newcite{habash2010introduction}. 
	
	For each token, we add a part of speech feature. For word form (non-clitic) tokens, we use the part of speech (POS) feature produced by the morphological analyzer. We consider the surface word and the lemma for representing the word form. For the clitics that were split off, we use a detailed POS feature that is also extracted from the output of the analyzer 
	and can take such forms as \textit{DET} for \textit{Al+} or \textit{poss\_pron\_3MP} for third person masculine possessive pronouns.   
	Table \ref{table3} shows the words and part of speech for the input sentence \textit{<fAstqblwhA>} \textit{`so they welcomed her'} \textit{fa-istaqbalu-ha}, using the lemma representation for the word form and the D3 tokenization scheme.
	
	These lexical and POS features are added to both our target model and sentiment model.

	\begin{table*}
		\begin{center}
				\begin{tabular}{ l  | l | l | l | l }
					\textbf{Word} & \textbf{English} & \textbf{Representation} & \textbf{POS} & \textbf{Token type} \\
					\textit{f}  & so & \textit{f+} &  conj &  clitic \\ 
					\textit{Astqblw}  & welcomed-they & \textit{isotaqobal\_1} &  verb & lemma \\ 
					\textit{hA} & her & \textit{+hA} & ivsuff\_do:3FS & clitic  \\ 
				\end{tabular}
		\end{center}
		\caption{\label{table3} {Example of morphological representation. The encoded features will be \textit{Representation} and \textit{POS}}. The POS for \textit{her} represents an object pronoun. The word form represented is the lemma.} 
	\end{table*}
	

	\subsection{Sentiment Features}
	The choice of sentiment lexicon is an important consideration when developing systems for new and/or low-resource languages. We consider three lexicons: (1) \textit{SIFAAT}, a manually constructed Arabic lexicon of 3982 adjectives \cite{abdul2011subjectivity}, (2) \textit{ArSenL}, an Arabic lexicon developed by linking English SentiWordNet with Arabic WordNet and an Arabic lexical database \cite{badaro2014large}, and (3) the English MPQA lexicon \cite{wilson2005recognizing}, where we look up words by matching on the English glosses produced by the morphological analyzer MADAMIRA. 
	
	For the target model, we add token-level binary features representing subjectivity, and for the sentiment model, we add both subjectivity  and polarity features. 
	
	
	We also add a feature specifying respectively the subjectivity or polarity of the parent word of the token in the dependency tree in the target or sentiment model. 
	
	\subsection{Syntactic Dependencies}
	
	We ran the CATiB (Columbia Arabic Treebank)  dependency parser \cite{shahrourimproving} on our data. CATiB uses a number of intuitive labels specifying the token's syntactic role: e.g \textit{SBJ, OBJ, MOD,} and \textit{IDF} for the Arabic \textit{idafa} construct (e.g \textit{<r'iys Al.hkwmT>} \textit{president of government}), as well as its part of speech role.
	In addition to the sentiment dependency features specifying the sentiment of parent words, we added dependency features specifying the syntactic role of the token in relation to its parent, and the path from the token to the parent, e.g \textit{nom\_obj\_vrb} or \textit{nom\_idf\_nom}, as well as the sentiment path from the token to the parent, e.g \textit{nom}(neutral)\textit{\_obj\_vrb}(negative) . 
	
	\subsection{Chunking and Named Entities}
	
	The morphological analyzer \textit{MADAMIRA} also produces base phrase chunks (BPC) and named entity tags (NER) for each token. We add features for these as well, based on the hypothesis that they will help define the spans for entity targets, whether they are named entities or any noun phrases.
	We refer to the sentiment and target models that utilize Arabic morphology, sentiment, syntactic relations and entity chunks as {\textbf{best-linguistic}.

\section{Word Clusters and Entity Semantics\label{clusters}}

Similar entities which occur in the context of the same topic or the same larger entity are likely to occur as targets alongside each other and to have similar sentiment expressed towards them. They may repeat frequently in a post even if they do not explicitly or lexically refer to the same person or object. For example, someone writing about American foreign policy may frequently refer to entities such as  \textit{\{the United States, America, Obama, the Americans, Westerners\}}. Such entities can cluster together semantically and it is likely that a person expressing positive or negative sentiment towards one of these entities may also express the same sentiment towards the other entities in this set.

Moreover, cluster features serve as a denser feature representation with a reduced feature space compared to Arabic lexical features. Such features can benefit the CRF where a limited amount of training data is available for target entities. 

To utilize the semantics of word clusters, we build word embedding vectors using the skip-gram method \cite{mikolov2013efficient} and cluster them using the K-Means algorithm \cite{macqueen1967some}, with Euclidean distance as a metric. Euclidean distance serves as a semantic similarity metric and has been commonly used as a distance-based measure for clustering word vectors.

The vectors are built on Arabic Wikipedia \footnote{https://dumps.wikimedia.org/arwiki/20160920/arwiki-20160920-pages-articles.xml.bz2} on a corpus of 137M words resulting in a vocabulary of 254K words. We preprocess the corpus by tokenizing (using the schemes described in section \ref{arabic}) and lemmatizing before building the word vectors.
We vary the number of clusters and use the clusters as binary features in our target and sentiment models.

\section{Experiments and Results\label{experiments}}

	\subsection{Experiments}
	
	\paragraph{Setup} 
	To build our sentiment and target models, we use CRF++ \cite{kudo2005crf++} to build linear-chain sequences. We use a context window of +/-2 for all features except the syntactic dependencies, where we use a window of +/-4 to better capture syntactic relations in the posts. For the sentiment model, we include the context of the previous predicted label, to avoid predicting consecutive tokens with opposite polarity. 
	
	We evaluate all our experiments on the development set which contains 116 posts and 442 targets, and present a final result with the best models on the unseen test. For the SentiWordNet-based lexicon ArSenL, we tune for the sentiment score threshold and use \textit{t=0.2}. We use Google's word2vec tool\footnote{https://github.com/dav/word2vec} for building and clustering word vectors with dimension 200. We vary the number of clusters $k$ between 10 (25K words/cluster) and 20K (12 words/cluster).
	\paragraph{Baselines} 
	 For evaluating the \textbf{predicted targets}, we follow work in English \cite{deng2015joint} and use the \textit{all-NP} baseline, where all nouns and noun phrases in the post are predicted as important targets. 
	
	For evaluating \textbf{sentiment towards targets}, we consider four baselines: the\textit{ majority} baseline which always predicts negative, and the \textit{lexicon} baseline evaluated in the case of each of our three lexicons: manually created, WordNet-based, and English-translated. The strong lexicon baseline splits the post into sentences or phrases by punctuation, finds the phrase that contains the predicted target, and returns positive if there are more positive words than negative words, and negative otherwise. These baselines are similar to the methods of previously published work for Arabic targeted sentiment \cite{al2015human,obaidat2015enhancing,abu2013identifying}.
	
	We run our pipelined models for all morphological representation schemes: \textit{surface word} (no token splits), \textit{lemma }(no clitics), lemma with ATB clitics (contain all token splits except \textit{Al+}), and lemma with D3 clitics (contains all token splits). We explore the effect of semantic word clusters in these scenarios. Finally we show our \textit{best-linguistic} (high-resource) model, and the resulting integration with word clusters.
	
	\subsection{Results}
	
	\begin{table*}
		\begin{center}
				\begin{tabular}{| l | l | l | l | l | l | l |l |}
					\hline
					& \multicolumn{3}{c|}{\textbf{Target}} & \multicolumn{4}{|c|}{\textbf{Sentiment}} \\ \cline{1-7} \hline
					\textbf{All-NP }& \textbf{Recall} & \textbf{Precision} & \textbf{F-score} & \textbf{F-pos} & \textbf{F-neg} & \textbf{Acc-sent} & \textbf{F-all} \\ \hline
					\textbf{Baseline1} Majority & 98.4  & 29.2 & 45 & 0 & 72.4 & 56.8 & 12.4   \\
					\textbf{Baseline2} ArSenL  & 98.4 & 29.2 & 45 & 50.6 & \bf 64.3 & 58.6 &  12.7 \\ 
					\textbf{Baseline3} SIFAAT & 98.4 & 29.2 & 45 & 61 & 58 &  59.5 & 13.1 \\ 
					\textbf{Baseline4} MPQA &98.4 & 29.2 & 45 & \bf 67 & 63.7 & \bf 65.4 & \bf 14.2 \bf \\  \hline  
				\end{tabular}
		\end{center}
		\caption{\label{table4} {Target and sentiment results using baselines; \textit{all-NP}} for targets and lexicons for sentiment.}
	\end{table*}
	
	\begin{table*}
		\begin{center}
				\begin{tabular}{| l | l | l | l | l | l | l |l |}
					\hline
					& \multicolumn{3}{c|}{\textbf{Target}} & \multicolumn{4}{|c|}{\textbf{Sentiment}} \\ \cline{1-7} \hline
					& \textbf{Recall} & \textbf{Precision} & \textbf{F-score} & \textbf{F-pos} & \textbf{F-neg} & \textbf{Acc-sent} & \textbf{F-all} \\ \hline
					\textbf{Surface} + POS & 41 & \bf 60.6 & 48.9 & 62.2 & 73.6 & 68.9  & 32.6 \\  
					\textbf{Lemma} + POS & 48.2\textsuperscript{**} &  60.5 & 53.7\textsuperscript{*} & \bf 65.4 & \bf 77.6 & \bf 72.8  & 38.1\textsuperscript{**} \bf \\  
					\textbf{+ATB} tokens  & 52.4\textsuperscript{*} & 59.5 & 55.7  & 61.3 & 75.7& 70.1  & \bf 
					38.2 \\   
					\textbf{+D3} tokens  & \bf 59.6\textsuperscript{**} & 55.7\textsuperscript{*} & \bf 57.6 & 64.1 & 73 & 69.2 &  36.1 \\  \hline 
				\end{tabular}
		\end{center}
		\caption{\label{table5}  {Target and sentiment results using different morphological representations. All models use POS.}} 
	\end{table*}
	
	Tables \ref{table4}-\ref{table6} show the results. Target F-measure is calculated using the \textit{subset} metric (similar to metrics used by \newcite{yang2013joint}, \newcite{irsoy2014opinion}); if either the predicted or gold target tokens are a subset of the other, the match is counted when computing F-measure.  Overlapping matches that are not subsets do not count (e.g <mwqf m.sr> \textit{Egypt's position} and <mwqf AsrA'iyl> \textit{Israel's position} do not match.). For this task, in the case of multiple mentions of the same entity in the post, any mention will be considered correct if the subset matches\footnote{We have also computed the performance for mention-overlap; the difference in target F-measure is 2 points and consistent across the different systems.}  (e.g  if <fls.tyn> \textit{Palestine} is a gold target, and <dwlT fls.tyn> \textit{state of Palestine} is predicted at a different position in the post, it is still correct). This evaluation is driven from the sentiment summarization perspective: we want to predict the overall opinion in the post towards an entity.
	
	\textit{F-pos}, \textit{F-neg}, and 	\textit{Acc-sent}  show the performance of the sentiment model on only the \textit{correctly predicted} targets\footnote{We exclude targets with ambiguous sentiment whose polarity was not agreed on by the annotators.}. Since the target and sentiment models are trained separately, this is meant to give an idea of how the sentiment model would perform in standalone mode, if targets were already provided.
	
	\textit{F-all} shows the overall F-measure showing the performance of correctly predicted targets with correct sentiment compared to the total number of polar targets. This evaluates the end-to-end scenario of both important target and sentiment prediction.
	
	Best results are shown in bold. Significance thresholds are calculated for the best performing systems (Tables \ref{table5}-\ref{table6}) using the approximate randomization test \cite{yeh2000more} for target recall, precision, F-measure, \textit{Acc-sent} and \textit{F-all}.  Significance over the method in the previous row is indicated by \textsuperscript{*} ($p<0.05$),\textsuperscript{**} ($p<0.005)$,\textsuperscript{**} ($p<0.0005$). A confidence interval of almost four F-measure points is required to obtain $p<0.05$. Our dataset is small; nonetheless we get significant results.

	
	\paragraph{Comparing Sentiment Lexicons}
	
	Table \ref{table4} shows the results comparing the different baselines. All targets are retrieved using \textit{all-NP}; sentiment is determined using the lexical baselines.
	As expected, the \textit{all-NP} baseline shows near perfect recall and low precision in predicting important targets. We observe that the gloss-translated MPQA lexicon outperforms the two other Arabic lexicons among the sentiment baselines. 

	We believe that the hit rate of MPQA
	is higher than that of the smaller, manually-labeled SIFAAT, and it is more precise than the automatically generated WordNet-based lexicon ArSenL. 
	The performance of MPQA is, however, reliant on the availability of high-quality English glosses. We found MPQA to consistently outperform in the model results, so in our \textit{best-linguistic} models, we only show results using the MPQA lexicon.
	
	\paragraph{Comparing Morphology Representations}
	
	Looking at table \ref{table5}, we can see that using the lemma representation easily outperforms the sparser surface word, 
	and that adding tokenized clitics as separate tokens outperforms  representations which only use the word form. Moreover, upon using the \textbf{D3} decliticization method, we observe a significant increase in recall of targets over the \textbf{ATB} representation. 
	This shows that the presence of the Arabic definite article <Al> \textit{Al+} is an important indicator of a target entity; thus, even if an entity is not named, \textit{Al+} indicates that it is a \textbf{known} entity and is likely more salient.
	
	The more tokens are split off, the more targets are recalled, although this comes at the cost of a decrease in sentiment performance, where the lemma representation has the highest sentiment score and the D3 representation has the lowest after surface word. We believe the addition of extra tokens in the sequence (which are function words and have not much bearing on semantics) generates noise with respect to the sentiment model.
	  All models significantly improve the baselines on  F-measure; for \textit{Acc-sent}, the surface word CRF does not significantly outperform the MPQA baseline.
	
	\paragraph{Effect of Word Clusters}
	Figures \ref{fig2} - \ref{fig5} show the performance of different morphological representations when varying the number of word vector clusters \textit{k}. (Higher \textit{k} means more clusters and fewer entities per semantic cluster.) 
	Adding cluster features tends to further boost the recall of important targets for all morphological schemes, while more or less maintaining precision. The difference in different schemes is consistent with the results of Table \ref{table5}; the D3 representation maintains the highest recall of targets, while the opposite is true for identifying sentiment towards the targets. The ATB representation shows the best overall F-measure, peaking at 41.5 using \textit{k}=250 (compare with 38.2 using no clusters); however, it recalls much fewer targets than the D3 representation.
	
	
	The effect of clusters on sentiment is less clear; it seems to benefit the D3 and ATB schemes more than lemma (significant boosts in sentiment accuracy). The improvements in F-measure and \textit{F-all} observed by using the best value of \textit{k} is statistically significant for all schemes (k=10 for lemma, k=250 for lemma+ATB, k=500 for lemma+D3, with \textit{F-all} values of 40.7, 41.5, and 39.1 respectively). In general, the cluster performances tend to peak at a certain value of \textit{k} which balances the reduced sparsity of the model (fewer clusters)  with the semantic closeness of entities within a cluster (more clusters). 

		\begin{figure}[ht]
			\centering
			\includegraphics[width=0.4\textwidth]{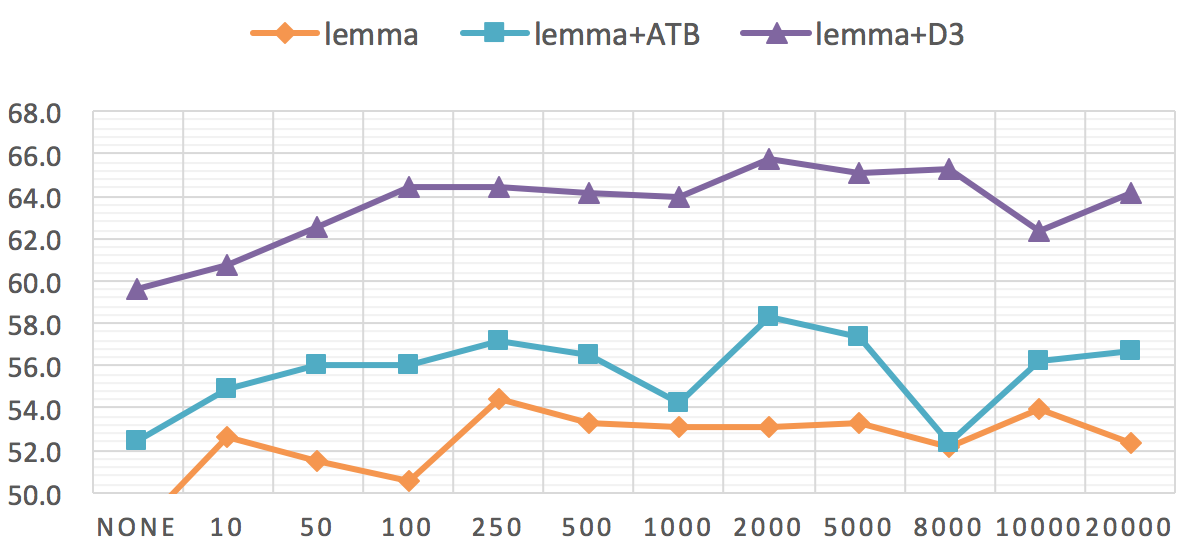}
			\caption{\label{fig2} Target recall vs clusters. }
	  	\end{figure}
		\begin{figure}[ht]
			\centering
			\includegraphics[width=0.4\textwidth]{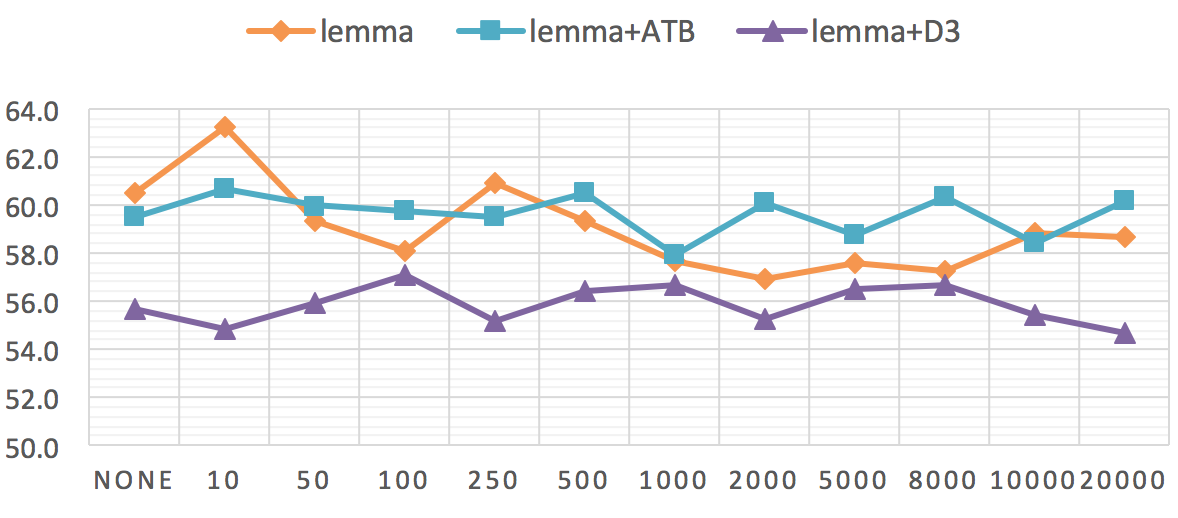}
			\caption{\label{fig3} Target precision vs clusters. }
		\end{figure}
		\begin{figure}[ht]
			\centering
			\includegraphics[width=0.4\textwidth]{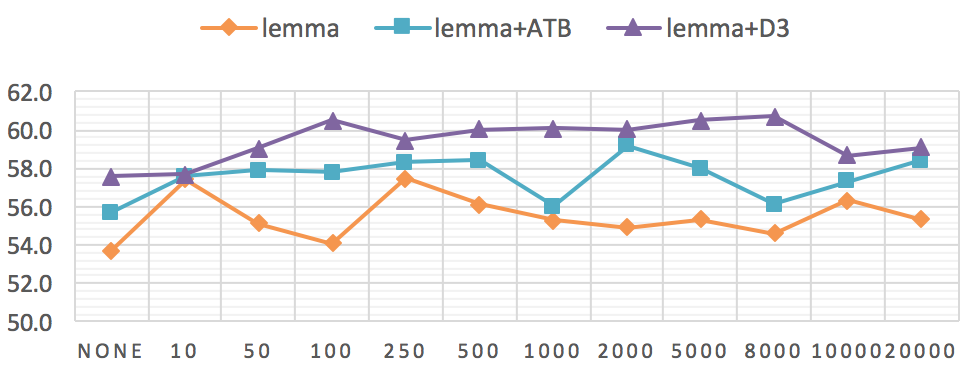}
			\caption{\label{fig4} Target F-score vs clusters. }
		\end{figure}
		\begin{figure}[ht]
			\centering
			\includegraphics[width=0.4\textwidth]{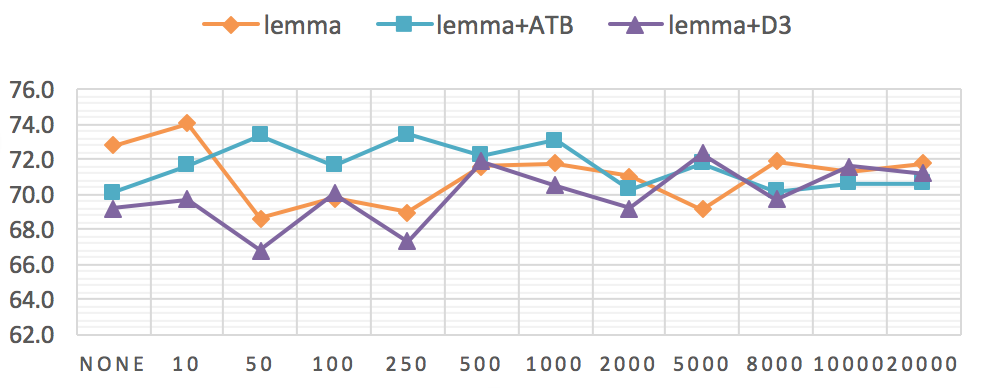}
			\caption{\label{fig5} Sentiment accuracy vs clusters. }
		\end{figure}
	   	\begin{figure}[ht]
			\centering
			\includegraphics[width=0.4\textwidth]{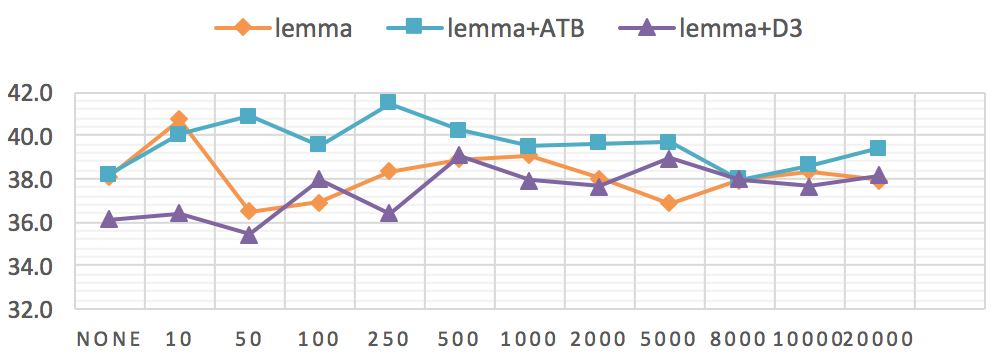}
			\caption{\label{fig6} Overall F-score vs clusters. }
		\end{figure}
	

	\paragraph{Performance of Best Linguistic Model}
	
		\begin{table*}
			\begin{center}
				\begin{tabular}{| l | l | l | l | l | l | l |l |}
					\hline
					& \multicolumn{3}{c|}{\textbf{Target}} & \multicolumn{4}{|c|}{\textbf{Sentiment}} \\ \cline{1-7} \hline
					& \textbf{Recall} & \textbf{Precision} & \textbf{F-score} & \textbf{F-pos} & \textbf{F-neg} & \textbf{Acc-sent} & \textbf{F-all} \\ \hline
					\textbf{best-linguistic}-ATB & 53 &  \bf 62.1 & 57.2  & 68.6  &  79.4 & 75.1 & 40.7   \\ 
					\textbf{best-linguistic}-D3  & 64.2\textsuperscript{***} & 58.8 & 61.4\textsuperscript{*} & 62.7 & 75.6 & 70.5\textsuperscript{*} &  39.1 \\ 
					\textbf{best-linguistic}-D3+ATB  & 63.7 & 58.8 &  61.4 & 67.7 & \bf 80 & 75.4\textsuperscript{***} & 43.1\textsuperscript{***} \\ 
					\textbf{best-linguistic}+clusters  &\bf 66.2 & 57.8 & \bf 61.8 & \bf{70} & \bf 80 & \bf{76} & \bf 44.2 \\ \hline
				\end{tabular}
			\end{center}
			\caption{\label{table6} {Performance of best linguistic model}}
		\end{table*}
	
	Table \ref{table6} shows the performance of our \textit{best-linguistic} model, which in addition to the word form and part of speech, contains named entity and base phrase chunks, the syntactic dependency features, and the sentiment lexicon features. The best linguistic model is run using both \textbf{ATB} and \textbf{D3} tokenization schemes, and then using a combined \textbf{ATB+D3} scheme where we use D3 for the target model and remove the extra clitics before piping in the output to the sentiment model. This combined scheme results in the best results overall: F-score of 61.4 for targets, accuracy of 75.4 for sentiment and overall F-measure of 43.1. 
	
	
	Adding the richer linguistic resources results in both improved target precision, recall, and sentiment scores, with  F-measure for positive targets reaching 67.7 for positive targets and 80 for negative targets. Performance exceeds that of the simpler models which use only POS and word clusters, but it is worth noting that using only the basic model with the word clusters can achieve significant boosts in recall and F-measure bringing it closer to the rich linguistic model.
	
	The last row shows the best linguistic model  \textbf{D3+ATB} combined with the clusters (best result for \textit{k}=8000, or about 30 words per cluster). Adding the clusters improves target and F-measure scores, although this result is not statistically significant. 
	We observe that it becomes more difficult to improve on the rich linguistic model using word clusters, which are more beneficial for low resource scenarios.

	
	

	Our results are comparable to published work for most similar tasks in English: e.g \newcite{yang2013joint} who reported target subset F-measure of \textasciitilde{}65, \newcite{pontiki2014semeval} where best performing SemEval systems reported 70-80\% for sentiment given defined aspects, and \cite{mitchell2013open,deng2015joint} for overall F-measure; we note that our tasks differ as described in section \ref{related}.
	
	
	
	
		\paragraph{Results on blind test}
		
		\begin{table}
			\begin{center}
				\setlength{\tabcolsep}{3.5pt}
				\begin{tabular}{| l | l | l | l | l | l |}
					\hline
					& \multicolumn{3}{c|}{\textbf{Target}} & \multicolumn{2}{|c|}{\textbf{Sentiment}} \\ \cline{1-5} \hline
					& \textbf{R} & \textbf{P} & \textbf{F} & \textbf{Acc} & \textbf{F-all} \\ \hline
					Best-D3 & 63.7 & \bf 52.3 & \bf 57.4 &  69.4 &  35.4 \\ 
					Best-D3+ATB & 63.7 & 51.8 &  57.1  &  70.3  & 36.8 \\ 
					+clusters  & \bf  65.6 & 50.2  &  56.9 & \bf 73.6  &   \bf 38.1\\  \hline 
				\end{tabular}
			\end{center}
			\caption{\label{table7} {Target and sentiment results 
					on test data.}} 
		\end{table}
		
		Table \ref{table7} shows the results on unseen test data for \textit{best-linguistic} using \textbf{D3}, \textbf{ D3+ATB} and with clusters using \textit{k}=8000. The results are similar to what was observed in the development data.

	\subsection{Error Analysis}
	
	
	
	\begin{table*}
		\begin{center}
			\setlength{\tabcolsep}{3.5pt}
			\begin{small}
			\begin{tabular}{ | l | }
				\hline
					 {\textbf{Example 1}}  \\  
					 {\fontsize{8}{1} \selectfont Till when will \textbf{[the world]-} wait before it intervenes against these \textbf{[crimes against humanity]-} committed by this \textbf{[criminal bloody} }  \\
					 {\fontsize{8}{1} \selectfont  \textbf{regime]-} which will not stop doing that... because its presence has always been associated with oppression and murder and crime...} \\
					{\fontsize{8}{1} \selectfont But now it's time for it to disappear and descend into \textbf{[the trash of history]-}}. \\ \hline
				   \textbf{Output} \quad \textbf{the world:}neg   \quad  \textbf{crimes:}neg \quad    \textbf{criminal bloody regime:}neg  \quad  \textbf{the trash of history}:neg    \\ \hline \hline
				   
				   {\textbf{Example 2}} \\
				   	{\fontsize{8}{1} \selectfont \textbf{[Malaysia]+} is considered the most successful country in Eastern Asia, and its economic success has spread to other \textbf{[aspects of life}} \\
				  	{\fontsize{8}{1} \selectfont  \textbf{in Malaysia]+}, for its \textbf{[services to its citizens]+} have improved, and there has been an increase in  \textbf{[the quality of its health and  }} \\
				   	{\fontsize{8}{1} \selectfont \textbf{educational and social and financial and touristic services]+}, which has made it excellent for foreign investments.} \\ \hline
				   	\textbf{Output} \quad \textbf{Malaysia:}pos   \quad  \textbf{health:}pos \quad    \textbf{educational and social:}neg  \quad  \textbf{financial}:neg    \\ \hline 
					
		
				\end{tabular}
			\end{small}
		\end{center}
		\caption{\label{table8} {Good and bad examples of output by SMARTies. Gold annotations for targets are provided in the text with `-' or `+' reflecting negative and positive sentiment towards targets.}} 
	\end{table*}

	We analyzed the output of our best linguistic models on the development set, and observed the following kind of errors:
	\paragraph{Implicit Sentiment} This was the most common kind of error observed. Commenters frequently expressed complex subjective language without using sentiment words, often resorting to sarcasm, metaphor, and argumentative language.
	We also observed persistent errors where positive sentiment was identified towards an entity because of misleading polar words; e.g \textit{minds <Al`qwl>} was consistently predicted to be positive even though the post in question was using implicit language to express negative sentiment; the English gloss is \textit{brains}, which appears as a positive subjective word in the MPQA lexicon. The posts also contained cases of complex coreference where subjective statements were at long distances from the targets they discussed.
	\paragraph{Annotation Errors} Our models often correctly predicted targets with reasonable sentiment which were not marked as important targets by annotators; this points to the subjective nature of the task. 
	
	\paragraph{Sentiment lexicon misses}
	These errors resulted from mis-match between the sentiment of the English gloss and the intended Arabic meaning, leading to polar sentiment being missed.
	\paragraph{Primary Targets} The data contains multiple entity targets and not all are of equal importance. Out of the first 50 posts manually analyzed on the dev set, we found that in 38 out of 50 cases (76\%) the correct \textit{primary} targets were identified (the most important topical sentiment target(s) addressed by the post); in 4 cases, a target was predicted where the annotations contained no polar targets at all, and in the remaining cases the primary target was missed. Correct sentiment polarity was predicted for 31 out of the 38 correct targets (81.6\%).
	
	
	
	In general, our analysis showed that our system does well on posts where targets and subjective language are well formed, but that the important target identification task is difficult and made more complex by the long and repetitive nature of the posts. Table \ref{table8} shows two examples of the translated output of SMARTies, the first on more well-formed text and the second on text that is more difficult to parse.
	
	

\section{Conclusions}
	We presented a linguistically inspired system that can recognize important entity targets along with sentiment in opinionated posts in Arabic.
	The targets can be any type of entity or event, 
    and they are not known beforehand. 
	Both target and sentiment results significantly improve multiple lexical baselines and are comparable to previously published results in similar tasks for English, a similarly hard task. Our task is further complicated by the informal and very long sentences that are used in Arabic online posts.
	We showed that the choice of morphological representation significantly
	affects the performance of the target and sentiment models. This could shed light on further research in target-specific sentiment analysis for morphologically complex languages, an area little investigated previously.
	We also showed that the use of semantic clusters boosts performance for both target and sentiment identification. Furthermore, semantic clusters alone can achieve performance close to a more resource-rich linguistic model relying on syntax and sentiment lexicons, and would thus be a good approach for low-resource languages. Integrating different morphological preprocessing schemes along with clusters gives our best result.
	
	Our code and data will be made publicly available\footnote{www.cs.columbia.edu/\textasciitilde{}noura/Resources.html}. Future work will consider cross-lingual clusters and morphologically different  languages. 
	
\section*{Acknowledgments}

This work was supported in part by grant NPRP 6-716-1-138 from the Qatar National Research Fund, by DARPA DEFT grant FA8750-12-2-0347 and by DARPA LORELEI grant HR0011-15-2-0041. The views expressed are those of the authors and do not reflect the official policy or position of the Department of Defense or the U.S government. We thank anonymous reviewers for their helpful comments. We thank Yves Petinot for providing feedback on the paper. We thank Nizar Habash and Mona Diab for helpful discussions.

\bibliography{eacl2017}
\bibliographystyle{eacl2017}

\end{document}